\RequirePackage{etoolbox}
\documentclass[cameraready]{Interspeech}
\usepackage{xurl} % allows line breaks at arbitrary characters
\newcommand{\model}[1]{\texttt{\def\UrlFont{\ttfamily}\path{#1}}}
\title{TID System for Speech-COSER Challenge 2026: ASR Track}

\author[affiliation={1, 2}, orcid=0000-0002-7705-2250, correspondingauthor]{Fernando}{L\'opez}
\author[affiliation={1}, orcid=0000-0002-4507-4930]{Jordi}{Luque}

\address{
    $^1$ Telef\'onica Innovaci\'on Digital, Spain\\
    $^2$ Universidad Aut\'onoma de Madrid, Spain
}
\email{fernando.lopez@telefonica.com}

\keywords{rural Spanish, speech recognition, speaker diarization, LoRA pruning, curriculum learning, ROVER fusion}

\usepackage{booktabs}
\usepackage{multirow}
\usepackage{xcolor, comment}

\begin{document}

\maketitle

\begin{abstract}
We describe the system submitted by Telefónica Innovación Digital to the Speech-COSER Challenge 2026, which features spontaneous speech from elderly speakers recorded in rural Spain. For ASR, we perform CTC-based pseudo-forced alignment scores audio-text match quality to guide a two-stage LoRA curriculum, followed by post-hoc magnitude pruning of the LoRA matrices to enhance generalization. An ensemble of seven systems combined via ROVER achieves 36.25\% unnormalized and 22.17\% normalized WER.%\review{In both tracks, we perform source annotation refinement followed by system fusion.} For ASR, CTC-based pseudo-forced alignment scores audio-text match quality to guide a two-stage LoRA curriculum, followed by post-hoc magnitude pruning of the LoRA matrices to enhance generalization. An ensemble of seven systems combined via ROVER achieves 36.25\% unnormalized and 22.17\% normalized WER. \review{For diarization, data is pseudo-annotated via DiariZen, followed by early fusion across encoders [TODO] and clustering [TODO], resulting in 18.65\% DER and 72.67\% JER.}
\end{abstract}

\section{Introduction}

The Speech-COSER Challenge 2026 is built on COSER~\cite{fernandez2018coser}, the \emph{Corpus Oral y Sonoro del Espa\~nol Rural}, comprising semi-guided interviews recorded between 1990 and 2025 across 1{,}468 rural locations in Spain (mean speaker age: 74.2 years). The speech is spontaneous, dialectal, and features frequent interruptions, overlaps, and untagged code-switching between Spanish and regional languages (Catalan, Galician, or Basque). $\sim$ 34\,h of audio is provided for training and development.

%This paper describes our submissions to both the automatic speech recognition (ASR) and speaker diarization (SD) tracks.

This paper describes our submission to the automatic speech recognition (ASR), our contributions are:
(i) We compute text--audio alignment confidence using iterative CTC pseudo-forced alignment~\cite{lopez2022iterative} to drive data filtering and curriculum selection.
(ii) We introduce a two-stage Low-Rank Adaptation (LoRA)~\cite{hu2022lora} curriculum~\cite{bengio2009curriculum} that trains on all source data prior to refinement on the filtered subset.
(iii) We explore post-hoc magnitude pruning of the learned LoRA adapters as a regularization mechanism to enhance generalization in low-resource regimes.
(iv) We evaluate ROVER~\cite{fiscus1997post} fusion across seven systems trained with this recipe, detailing individual voter contributions.

%\review{[TODO: For SD, our contributions are]}

\section{Data and Annotation Refinement}
\label{sec:data}

\subsection{Challenge Data}
\label{sec:release}

The challenge distributes two subsets: 
(i) \emph{Pre-segmented}: 14{,}487 short clips ($<$30\,s) from 35 recordings with human-reviewed transcriptions, totalling 23\,h of audio. 
(ii) \emph{Long files}: 8 full-length recordings ($\sim$11.5\,h) annotated at the speaker-turn level (\texttt{[start][end][speaker]: text}). This subset is neither cleaned nor manually validated.

%\subsection{ASR: Text--Audio Alignment Score}
\subsection{Text--Audio Alignment Score}
\label{sec:alignment}

Manual validation confirmed that transcriptions generally match the audio. Nonetheless, discrepancies occur, including text assigned to adjacent segments, partial or incorrect transcripts, and drifted turn boundaries. Because full manual validation is impractical, we adopt an automatic validation method.

We apply the iterative pseudo-forced alignment procedure of~\cite{lopez2022iterative} with \model{omniASR-CTC-7B}. The method aligns reference transcripts against CTC posterior probabilities to compute $\mathrm{Segment\_Score}$, defined as the average log-probability along the alignment path. Low scores stem from acoustic mismatch, overlapping speech, transcription drift, or long sequences.

For pre-segmented clips, we retain the original boundaries, align each clip against its reference, and record the alignment score. For long files, alignment begins from the source timestamps, using CTC to refine speech boundaries.

\begin{figure}[t]
  \centering
  \includegraphics[width=\columnwidth]{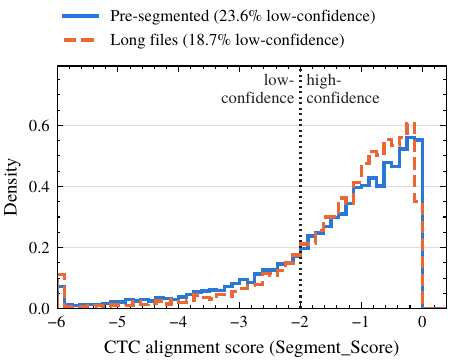}
  \caption{CTC alignment score distributions for pre-segmented clips and long-file turns. The vertical line marks the $-2.0$ threshold separating low- and high-confidence utterances.}
  \label{fig:alignment}
\end{figure}

Figure~\ref{fig:alignment} shows the resulting score distributions. Based on listening across score tiers, we set an empirical threshold at $-2.0$, discarding 23.6\% of pre-segmented clips and 18.7\% of long-file turns. While low scores reflect combined acoustic and alignment difficulty rather than transcription errors alone, we operationally denote retained utterances as \emph{high-confidence} and discarded ones as \emph{low-confidence}. The retained pool forms the training set for our second curriculum stage (Section~\ref{sec:curriculum}).

\textbf{Reference text cleaning.} COSER transcriptions include inline markers requiring normalization:
dialectal--standard pairs (\texttt{habr\'is=habr\'eis}) are mapped to the spoken form (\texttt{habr\'is}) to reflect the acoustics; overlapping speech tags (\texttt{[HS:ID text]}) are unwrapped to retain spoken words; and non-speech events or anonymization tags (\texttt{[RISAS]}, \texttt{[Anonim]}) are stripped.

%\subsection{SD: Pseudo-annotation}
%\label{sec:sd-data}

%\review{[TODO: pseudo-annotation of the diarization data with DiariZen.]}

\section{ASR System}
\label{sec:asr}

\subsection{Model and Two-Stage Curriculum}
\label{sec:curriculum}

Our base model is \model{Voxtral-Mini-3B-2507}~\cite{liu2025voxtral}, a large audio language model (LALM) pairing a Whisper-based audio encoder~\cite{radford2023robust} with a 3B-parameter language model (Ministral~3B backbone~\cite{liu2026ministral3}). We fine-tune it with LoRA on the attention and feed-forward projections of both encoder and decoder, as well as the audio-to-text projector, using $\alpha/r=2$, dropout 0.10, weight decay 0.05, 400 warm-up steps, and early stopping on development loss.

\textbf{Stage 1} trains on all training data with learning rate $1.5\times10^{-4}$. \textbf{Stage 2} resumes from the Stage~1 checkpoint on the high-confidence subset only at $3\times10^{-5}$ for up to two epochs. This two-stage schedule first captures broad domain acoustic properties before refining on cleaner audio--text alignments.

\subsection{Post-Hoc LoRA Pruning}
\label{sec:pruning}

\textbf{Motivation.} While training loss decreased steadily, validation metrics plateaued, indicating adapter overfitting due to scarce data: $\sim$30\,h from 36 recordings in Stage~1, and $\sim$23\,h following alignment filtering in Stage~2.

\textbf{Method.} After training, we keep the fraction $k$ of largest absolute value entries in each LoRA factor matrix ($A$ and $B$) across all adapted layers, zeroing out the remaining entries. The pruned adapter requires no architectural modifications or retraining; searching over $k$ incurs only decoding cost. We prune both stage endpoints and select $k$ on the internal validation split.

\textbf{Related work.} Magnitude pruning typically involves iterative compression followed by retraining~\cite{han2015learning}, as applied to self-supervised speech encoders in low-resource ASR by PARP~\cite{lai2021parp}. In parameter-efficient fine-tuning (PEFT), AdaLoRA~\cite{zhang2023adalora} dynamically prunes singular values during training; SparseAdapter~\cite{he2022sparseadapter} prunes adapter matrices at initialization prior to tuning; and S2-LoRA~\cite{liu2024s2lora} employs shared low-rank matrices with sparse rank coefficients. ComPEFT~\cite{yadav2025compeft} sparsifies PEFT weights post hoc, targeting parameter reduction. Post-hoc interventions also regularize out-of-distribution transfer: WiSE-FT~\cite{wortsman2022robust} interpolates fine-tuned and pre-trained weights, LASER~\cite{sharma2024laser} performs low-rank approximation on frozen matrices, while model-merging techniques such as TIES~\cite{yadav2023ties} and DARE~\cite{yu2024dare} trim or drop delta parameters. In contrast, our approach applies post-hoc unstructured magnitude pruning directly to the low-rank factor matrices ($A$ and $B$) as an inexpensive regularizer against in-domain overfitting.

\subsection{Decoding and Post-Processing}
\label{sec:decoding}

Because evaluation files are full-length recordings, audio is segmented prior to inference. We run Silero VAD~\cite{SileroVAD} (threshold 0.3, speech gaps $<$500\,ms merged) and pack contiguous speech into windows of up to 25\,s. Specifically, windows preserve natural pauses within continuous spans without splicing separate segments, and cuts are placed at VAD activation minima within a boundary search margin~\cite{bain2023whisperx}.

Decoded window transcripts are concatenated per session and processed with a rule-based hallucination filter~\cite{lopez26b_interspeech} that: (i) detects character-level repetitions in words exceeding 20 characters and truncates them to their root token; (ii) collapses 3 or more consecutive repeated words into a single occurrence; and (iii) deduplicates 3 or more repeated multi-word phrases.

\subsection{Additional Voters and ROVER Fusion}
\label{sec:fusion}

As improvements from fine-tuning a single architecture diminished, we adopted late fusion across diverse model families. Candidate selection was guided by a zero-shot benchmark of eleven ASR models across five Iberian languages~\cite{lopez2026benchmarking}. On its Spanish subset, \model{SeamlessM4T-v2}~\cite{seamless2023} (9.1\% WER), \model{Qwen3-ASR-1.7B}~\cite{shi2026qwen3} (10.0\%), and the 1B variant of \model{Omnilingual-ASR}~\cite{omnilingual2025omnilingual} (10.2\%) outperformed \model{Voxtral-Mini-3B} (12.5\%). We selected these three: scaling to \model{omniASR-LLM-7B-v2}, and included \model{Canary-1B-v2}~\cite{sekoyan2025canary} from the same benchmark. Finally, \model{Qwen3-Omni-30B-A3B}~\cite{xu2025qwen3} was added as the top-performing model on ESCUCHA~\cite{lopez2026escucha}, a benchmark for non-normative and multi-accent Spanish speech understanding where it achieved 74.40\% accuracy versus 49.20\% for \model{Voxtral-Mini-3B}.

All models were fine-tuned with the recipe from Section~\ref{sec:curriculum}: identical splits, the two-stage curriculum, and a pruning sweep across both endpoints. Target modules differed in two cases: for \model{omniASR-LLM-7B-v2}, LoRA was applied to the top 16 encoder layers and the entire decoder; for \model{Qwen3-Omni-30B-A3B}, expert weights were frozen in FP8, restricting LoRA to audio encoder and decoder attention projections. For each architecture, the curriculum checkpoint and pruning ratio $k$ were selected based on marginal contribution to ensemble performance on the internal test split rather than individual WER.

During evaluation, all models decode identical VAD segments, enabling window-level alignment. We combine hypotheses using ROVER~\cite{fiscus1997post}, aligning candidates to a pivot (the best single system) and resolving slots via majority vote, breaking ties in favor of the pivot~\cite{umesh2022bcn2brno}. Punctuation and capitalization are voted independently from word tokens: once normalized lexical identities are established, casing and punctuation are determined across all systems aligning a token to that slot. For official submissions, raw window hypotheses from all voters are fused before applying the hallucination filter (Section~\ref{sec:decoding}) to the composite transcript, whereas internal experiments fuse post-filtered hypotheses.

%\section{SD System}
%\label{sec:diarization}

% TODO(diarization team): method only here; numbers go to Section~\ref{sec:res-sd}.
%\review{[TODO: early fusion across encoders and clustering.]}

\section{Experimental Setup}
\label{sec:setup}

%\subsection{ASR Data Splits}
\subsection{Data Splits}
\label{sec:asr-splits}

We partition the data by recording to prevent speaker leakage. The development set comprises 4 pre-segmented recordings (Valencia, L\'erida, Murcia, Canarias; 1{,}492 segments), and the internal test set comprises 3 (Asturias, Arag\'on, Pa\'{\i}s Vasco; 720 segments). All remaining data, including all long files, serve as training material: 17{,}473 segments ($\approx$30\,h) in Stage~1 and the high-confidence subset of 14{,}328 segments ($\approx$23\,h) in Stage~2. Hyperparameter tuning, pruning sweeps, and fusion candidates are evaluated exclusively on the 720-segment test split.

\begin{figure*}[t!]
  \centering
  \includegraphics[width=\textwidth]{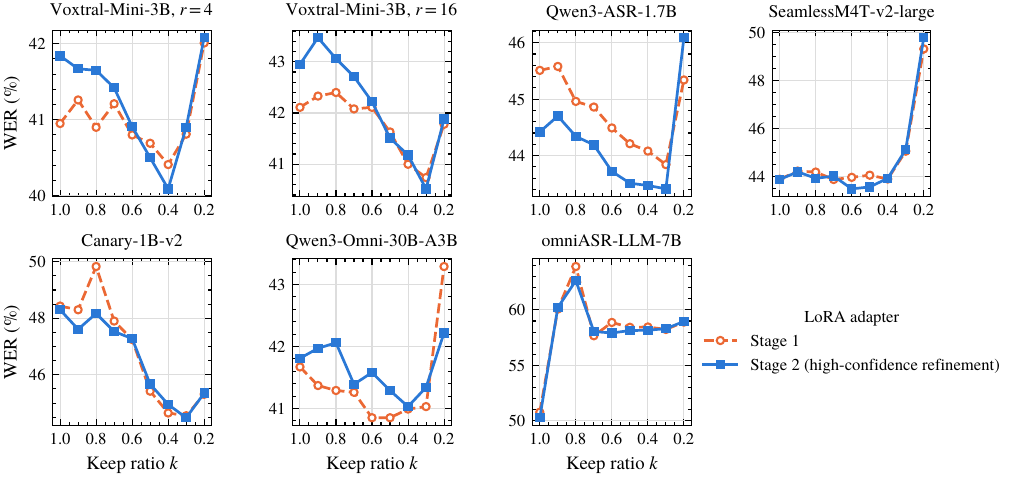}
  \caption{Post-hoc LoRA magnitude pruning on the internal test set. Curves show two stages WER across keep ratios $k \in [0.2, 1.0]$.}
  \label{fig:pruning}
\end{figure*}

\begin{table}[t]
  \caption{Single-system results for \texttt{Voxtral-Mini-3B} across internal (720 segments) and official evaluation sets.}
  \label{tab:single}
  \centering
  \setlength{\tabcolsep}{4pt}
  \footnotesize
  \begin{tabular}{lcccc}
    \toprule
    & \multicolumn{2}{c}{\textbf{Internal}} & \multicolumn{2}{c}{\textbf{Official}} \\
    \textbf{System} & WER & WER$_\mathrm{norm}$ & WER & WER$_\mathrm{norm}$ \\
    \midrule
    Zero-shot & 49.20 & 35.63 & -- & -- \\
    Initial LoRA submission & 47.63 & 33.34 & 45.47 & 31.61 \\
    Best single system & \textbf{40.09} & \textbf{26.22} & \textbf{38.22} & \textbf{24.06} \\
    \bottomrule
  \end{tabular}
\end{table}

%\subsection{ASR Evaluation Metrics}
\subsection{Evaluation Metrics}
\label{sec:asr-metrics}

The primary competition metric is unnormalized Word Error Rate (WER), which compares hypotheses and references verbatim, counting casing and punctuation discrepancies as errors. The secondary metric is WER$_\mathrm{norm}$, which normalizes by lowercasing text and stripping punctuation and anonymization tags. Both metrics are computed using MeetEval~\cite{vonneumann2023meeteval}, aggregating errors across the 720 internal test segments. Official results are scored by the challenge organizers on full-session transcripts.

%\subsection{SD Data Splits}
%\label{sec:sd-splits}
%\review{[TODO: SD data splits: leave one out]}

%\subsection{SD Evaluation Metrics}
%\label{sec:sd-metrics}
%\review{[TODO: SD metrics (DER, JER) and their settings, e.g., collar.]}

\section{Results}
\label{sec:results}

\begin{table*}[t]
  \caption{Incremental ROVER ensemble. Voters are accumulated top to bottom. \emph{Individual system}: internal test performance zero-shot and fine-tuned (S1/S2: curriculum stage; $k$: LoRA keep ratio). \emph{Ensemble}: cumulative pool evaluated internally and on official submissions; $\Delta$ denotes official improvement over the preceding submission (row~4 relative to single-system row~1). All values are \%.}
  \label{tab:fusion}
  \centering
  \setlength{\tabcolsep}{4.5pt}
  \resizebox{\textwidth}{!}{%
  \begin{tabular}{rll cc cc cc cccc}
    \toprule
    & & & \multicolumn{4}{c}{\textbf{Individual system (internal)}} & \multicolumn{6}{c}{\textbf{Ensemble after adding this voter}} \\
    \cmidrule(lr){4-7} \cmidrule(lr){8-13}
    & & & \multicolumn{2}{c}{Zero-shot} & \multicolumn{2}{c}{Fine-tuned} & \multicolumn{2}{c}{Internal} & \multicolumn{4}{c}{Official} \\
    \cmidrule(lr){4-5} \cmidrule(lr){6-7} \cmidrule(lr){8-9} \cmidrule(lr){10-13}
    \# & \textbf{Voter} & \textbf{Checkpoint} & WER & WER$_\mathrm{norm}$ & WER & WER$_\mathrm{norm}$ & WER & WER$_\mathrm{norm}$ & WER & $\Delta$ & WER$_\mathrm{norm}$ & $\Delta$ \\
    \midrule
    1 & \model{Voxtral-Mini-3B}, $r{=}4$  & S2, $k{=}0.40$ & 49.20 & 35.63 & 40.09 & 26.22 & 40.09 & 26.22 & 38.22 & --    & 24.06 & -- \\
    2 & \model{Voxtral-Mini-3B}, $r{=}16$ & S2, $k{=}0.35$ & 49.20 & 35.63 & 40.52 & 26.62 & --    & --    & --    & --    & --    & -- \\
    3 & \model{Qwen3-ASR-1.7B}            & S2, $k{=}0.35$ & 58.54 & 39.37 & 43.38 & 28.60 & 39.01 & 25.29 & --    & --    & --    & -- \\
    4 & \model{omniASR-LLM-7B-v2}         & S2, unpruned   & 59.05 & 37.32 & 50.28 & 34.10 & 38.99 & 25.03 & 37.32 & $-0.90$ & 23.07 & $-0.99$ \\
    5 & \model{SeamlessM4T-v2}            & S2, $k{=}0.40$ & 62.73 & 46.12 & 43.93 & 30.02 & 38.06 & 24.35 & 36.88 & $-0.44$ & 22.72 & $-0.35$ \\
    6 & \model{Canary-1B-v2}              & S1, $k{=}0.35$ & 50.61 & 36.80 & 44.20 & 29.79 & 37.78 & 24.14 & 36.71 & $-0.17$ & 22.50 & $-0.22$ \\
    7 & \model{Qwen3-Omni-30B-A3B}        & S1, $k{=}0.35$ & 48.55 & 35.12 & 40.75 & 26.50 & \textbf{37.45} & \textbf{23.89} & \textbf{36.25} & $-0.46$ & \textbf{22.17} & $-0.33$ \\
    \midrule
    \multicolumn{7}{r}{\emph{Total gain of the 7-voter ensemble over the best single system}} & & & & $-1.97$ & & $-1.89$ \\
    \bottomrule
  \end{tabular}}
\end{table*}

\subsection{Single System}
\label{sec:res-single}

Table~\ref{tab:single} reports the zero-shot baseline, our initial LoRA submission, and our best single system. Zero-shot \model{Voxtral-Mini-3B} achieves 49.20\% unnormalized WER internally. Successive training and decoding optimizations (text cleaning, curriculum, LoRA pruning, and windowing) reduced official test WER from 45.47\% to 38.22\%.

\textbf{Curriculum and pruning.} Unpruned Stage~2 checkpoints perform on par with or worse than Stage~1, indicating that refining exclusively on high-confidence data causes adapter overfitting. However, post-hoc magnitude pruning reverses this trend. As shown in Figure~\ref{fig:pruning}, WER exhibits a clear optimum as $k$ decreases across nearly all configurations. For \model{Voxtral-Mini-3B}, pruning improves the Stage~2 adapter by 1.75\,pp ($r{=}4$, $k{=}0.40$) and 2.43\,pp ($r{=}16$, $k{=}0.30$), with performance degrading sharply below $k{=}0.20$. Pruned Stage~2 models consistently surpass Stage~1, validating the two-stage curriculum when regularized. The exceptions are \model{omniASR-LLM-7B-v2} (underconverged with the same recipe) and \model{Qwen3-Omni-30B-A3B}, where LoRA was restricted to attention layers with frozen expert weights.

\textbf{Decoding window.} Window length was the primary decoding driver: expanding from 10\,s to 25\,s improved official WER by 3.15\,pp by reducing mid-word segment boundary cuts from 18.6\% to 2.9\%. This effect is unobservable on internal test segments, which are pre-segmented.

%\subsection{ASR Fusion}
\subsection{System Fusion}
\label{sec:res-fusion}

\textbf{Voters.} General-domain benchmark performance~\cite{lopez2026benchmarking} did not directly translate to elderly rural speech: \model{SeamlessM4T-v2} yielded the poorest zero-shot baseline on our internal test (62.73\% WER, Table~\ref{tab:fusion}), yet proved to be one of the most effective ensemble voters post-adaptation. Furthermore, the tuning recipe behaved differently across architectures (Figure~\ref{fig:pruning}). Pruning benefits most models: sharply for \model{Canary-1B-v2} (up to 3.9\,pp) and \model{Voxtral-Mini-3B}, moderately for \model{Qwen3-ASR-1.7B} and \model{Qwen3-Omni-30B-A3B}, and minimally for \model{SeamlessM4T-v2}, whose curve remains flat before degradation. Conversely, pruning degraded \model{omniASR-LLM-7B-v2} across all ratios ($+7$ to $+13$\,pp) due to incomplete training convergence. Optimal curriculum endpoints also diverged: post-pruning, Stage~2 was superior for \model{Voxtral-Mini-3B}, \model{Qwen3-ASR-1.7B}, and \model{SeamlessM4T-v2}; both stages tied for \model{Canary-1B-v2}; and Stage~1 performed best for \model{Qwen3-Omni-30B-A3B}.

\textbf{Fusion.} Independent voting for punctuation and capitalization gained 0.48\,pp WER over binding them to lexical token winners. Diversity outweighed individual accuracy: incorporating voters up to 10\,pp worse than the top single system consistently yielded gains. Odd-sized pools (3, 5, or 7 voters) also mitigated tie resolutions inherent to even combinations. Table~\ref{tab:fusion} outlines cumulative ensemble progression. The initial 4-voter pool (\model{Voxtral-Mini-3B} $r{\in}\{4, 16\}$, \model{Qwen3-ASR-1.7B}, and \model{omniASR-LLM-7B-v2}) improved official WER by 0.90\,pp over the best single system; successive additions of \model{SeamlessM4T-v2}, \model{Canary-1B-v2}, and \model{Qwen3-Omni-30B-A3B} contributed 0.44, 0.17, and 0.46\,pp, respectively. The final 7-voter ensemble attained 36.25\% WER, and 22.17\% normalized WER (a 1.97\,pp net gain over our best single model). Notably, internal validation gains accurately mirrored the trajectory and direction of official evaluation results within a factor of two, showing far higher reliability than decoding window tuning.

%\subsection{SD results}
%\label{sec:res-sd}

%\review{[TODO: internal and official DER/JER (official: 18.65\% DER, 72.67\% JER).]}

\section{Limitations}
\label{sec:limitations}

%For ASR:
Several limitations apply:
(i) Alignment filtering relies on CTC posteriors from a single model (\model{omniASR-CTC-7B}); errors or biases in the alignment model directly propagate into the curriculum split.
(ii) Post-hoc LoRA pruning requires an empirical sweep, and pruning rates do not transfer uniformly across distinct model families. (iii) The training recipe used for all models was optimized only for \model{Voxtral-Mini-3B-2507}, bigger improvements are expected from model-specific optimization.
%\review{[TODO: Limitations for diarization.]}

\section{Conclusions}
\label{sec:conclusions}

%For ASR, 
We scored text--audio alignment to guide curriculum data filtering. Pairing these two-stage curriculum with post-hoc magnitude pruning of LoRA adapters yielded our best single system. Pruning requires no retraining, incurs negligible post-processing cost, and improved five of seven evaluated systems, though optimal retention ratios and curriculum endpoints remain architecture-dependent. Finally, late fusion via ROVER across seven diverse models trained with this unified recipe delivered our most consistent improvements, reducing official test WER from 38.22\% to 36.25\% (22.17\% normalized WER). %\review{[TODO: one sentence on the diarization conclusions.]}

%\bibliographystyle{IEEEtran}
%\bibliography{mybib}


\end{document}